%% file: main.tex
\documentclass[runningheads]{llncs}
\AtBeginDocument{\hypersetup{pdfborder={0 0 0}}}
\usepackage[T1]{fontenc}
\usepackage{caption}
\DeclareCaptionFont{9pt}{\fontsize{9pt}{10pt}\selectfont}
\usepackage{hyperref}
\usepackage{graphicx}
\usepackage{subfiles}
\usepackage{amssymb}

\input{_constants}
\usepackage{comment}
\usepackage{booktabs} 
\usepackage{amsmath}
\usepackage{cleveref}
\usepackage{subfig}

\author{Pierrick Chatillon\inst{1,2} \and
Yann Gousseau\inst{1} \and
Sidonie Lefebvre \inst{2}}
\institute{LTCI, Télécom Paris, IP Paris, 
19 place Marguerite Perey 91120 Palaiseau, France \email{\{name.surname\}@telecom-paris.fr} \and
DOTA \& LMA2S, ONERA, Université Paris Saclay 
F-91123 Palaiseau, France
\email{\{name.surname\}@onera.fr}}

\begin{document}
\title{A geometrically aware auto-encoder for multi-texture synthesis}
%

\maketitle              
\begin{abstract}
We propose an auto-encoder architecture for multi-texture synthesis. The approach relies on both a compact encoder accounting for second order neural statistics and a generator incorporating adaptive periodic content. Images are embedded in a compact and geometrically consistent latent space, where the texture representation and its spatial organisation are disentangled. Texture synthesis and interpolation tasks can be performed directly from these latent codes. 
Our experiments demonstrate that our model outperforms state-of-the-art feed-forward methods in terms of visual quality and various texture related metrics. The code is available online.

\end{abstract}

\subfile{01_intro}

\vspace{-0.5em}
\subfile{02_related}

\vspace{-0.5em}
\subfile{03_method}
\subfile{04_experiments}

\vspace{-.5em}
\subfile{10_conclusion}
\vspace{-.5em}
\section{Acknowledgements}
This work was supported by the Defence Innovation Agency and the project MISTIC (ANR-19-CE40-005).

{\small
\bibliographystyle{splncs04}
\bibliography{main.bib}
}
\end{document}

%% file: _constants.tex
\newif\ifreview 
\newif\ifarxiv 
\newif\ifcamera 
\newif\ifrebuttal 

%% file: 01_intro.tex
\section{Introduction}
\label{sec:intro}

 Texture synthesis, that is the process of synthesizing new image samples from a given exemplar, has experienced a clear breakthrough with the work of Gatys et al. \cite{gatys2015texture}. Inheriting ideas from wavelet-based methods, it was first proposed in this work to synthesize textures by constraining second order statistics of the responses to classical classification neural networks. While this approach outperformed existing ones in term of visual fidelity, it relies on a relatively heavy optimization procedure that has to be carried out for each new exemplar. Therefore, methods have been proposed to perform synthesis using generative neural networks, such as networks learned for each new exemplar \cite{ulyanovtexture} or GANs \cite{SGAN}. More recently, such approaches have been extended to perform the synthesis from a whole set of textures, by using adaptive normalizations of generative networks \cite{WCT} \cite{texton}.
 \let\thefootnote\relax\footnotetext{Code available at: \href{https://github.com/PierrickCh/TextureAutoEncoder}{https://github.com/PierrickCh/TextureAutoEncoder}}

The present work is in the continuation of these approaches and introduces an auto-encoder architecture enabling one to synthesize arbitrary textures from a compact latent representation. The autoencoder architecture avoids the training instability inherent to GAN-based approaches and is naturally suited to synthesis and editing tasks. 
The encoding step yields a latent code adapted to an input texture, from which arbitrary samples can be synthesized. The resulting latent space is spatially agnostic, or in other words the network treats in the same way different translations of a given input, which is made possible by carefully designing both the encoder and the generator. 
Moreover, the proposed architecture  deals with both stochastic and structured, periodic-like textures. This is achieved thanks to the inclusion of sine waves in the design of the generator. Contrarily to what is done in the literature, we deal with these periodic components in a way that allows arbitrary rotations and scalings of the exemplar textures, a property that is clearly desirable in order to achieve a generic representation of textures. To the best of our knowledge, the resulting architecture is the first approach proposing a generic latent representation for textures, including an encoder to embed texture images, from which new samples can be synthesized or in which editing operations can be performed.

%% file: 02_related.tex
\section{Related Work}
\label{sec:related}
\subsection{Periodic Texture Synthesis}
Synthesizing periodic textures and textures with long range dependency is a challenging aspect of texture synthesis. 
In order to enable the synthesis of multiple stationary periodic textures, the authors of \cite{PSGAN} improved the GAN-based synthesis method from \cite{SGAN} by providing additional long range periodic information to the generator. This concept has proven efficient and was reused in \cite{texton}, coupled with the structure of a StyleGAN2 \cite{StyleGAN2} network. Another approach is proposed in \cite{NeuralTextureGonthier} by imposing some constraints on the Fourier power spectrum. 
\vspace{-1em}
\subsection{Universality and latent representations of textures}
Beyond the classical problem of exemplar-based texture synthesis, where the goal is to generate samples from a single input texture, a difficult problem is to develop a neural architecture that has the ability to synthesize arbitrary texture inputs, possibly not seen during the training phase. Incidentally, this also raises the question of how to efficiently encode textures, e.g. through latent spaces. The authors of \cite{WCT} were the first to propose such an approach, leveraging the concept of adaptive instance normalization \cite{adain} to provide a flexible and universal style transfer and texture synthesis method, where each texture is represented by several thousands normalizing parameters related to VGG19 features. To allow for user control \cite{user-controllable} proposes a modification of PSGAN \cite{PSGAN} 
by introducing a convolutional encoder. They focus on very small latent spaces 
well suited to the representation of patches from a given image. Authors from \cite{neural} also use a convolutional encoder to generate textures, with exemplar-specific geometric transformations applied to the noise input of the generator.
Building on attention mechanisms, similarity maps or Fourier representations, some works \cite{Uatt}\cite{transposer} also tackle the problem of universal neural texture synthesis but suffer from verbatim copy of structures, such as those usually observed in patch-based approaches. 
The method from \cite{texton} also has the ability to synthesize a variety of textures, but relies on a latent space which needs inverting the generator to represent given inputs.

%% file: 03_method.tex
\section{Method}
\label{sec:method}

\subsection{Architecture overview}
Our texture synthesis framework relies on an auto-encoder structure, that we call in short \textbf{TAE}. As explained in the introduction, we wish to develop a latent space that is spatially agnostic, deals with simple geometric transforms and with periodic textures. 

The encoder $E$ is used to map an image to a latent texture representation space $\mathcal{W}$. From this representation, new samples with different spatial arrangements can be generated with the generator $G$. For this generator, we rely on the classical StyleGAN architecture \cite{StyleGAN2}. This architecture is augmented with spatially periodic information, namely sine maps as used in \cite{PSGAN}\cite{texton}. But unlike them, the periodic information is modulated using the latent representation and is thus adapted to each input sample. This adaptation of the periodic content is crucial and achieved by predicting scale and orientation parameters from the latent variable $w$.

\begin{figure}[t!]
    \centering
      \includegraphics[angle=0,width=1.\linewidth]{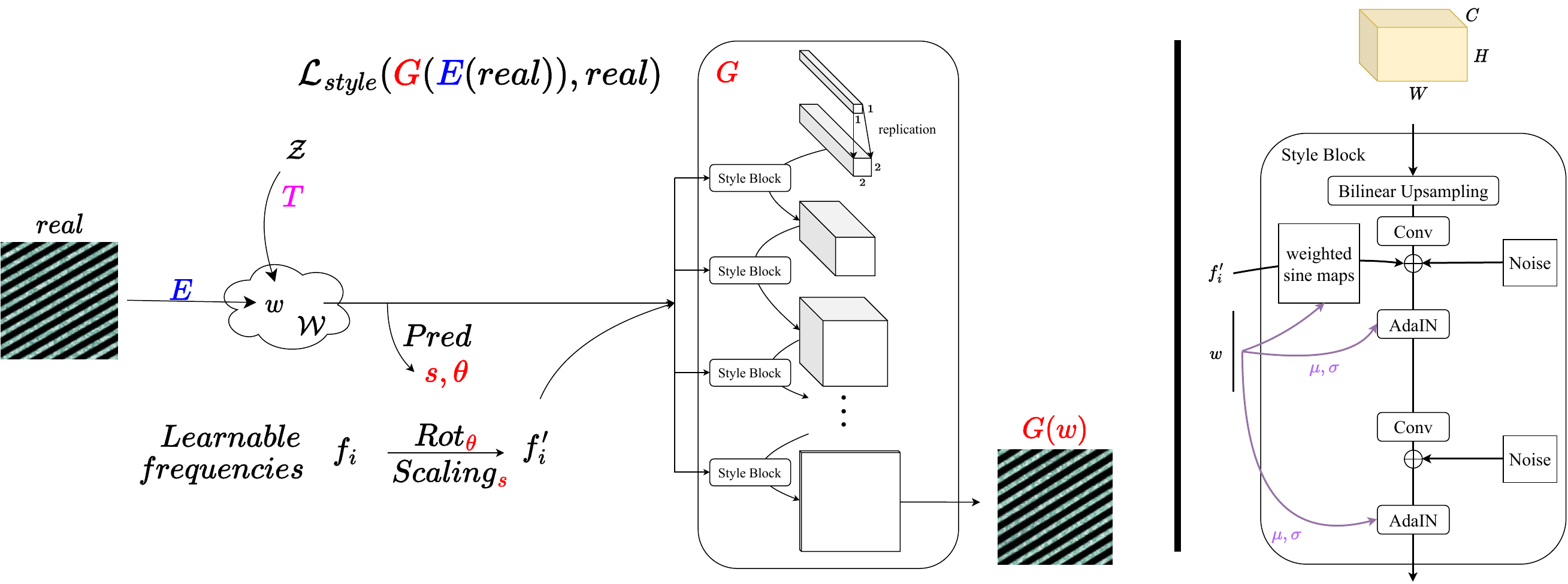}
      \caption{Left: Texture Auto-encoder architecture. Right: Modified style block}\,

      \label{TAE}
      \vspace*{-.2cm}
\end{figure}

The whole pipeline is mainly supervised by the texture reconstruction loss $\mathcal{L}_{style}$ \cite{gatys2015texture} between the input and its reconstruction by the texture auto-encoder.
Additionally, we introduce a network $T$ designed for direct texture sampling, enabling us to map a noise distribution onto our latent space $\mathcal{W}$ in a meaningful way. This allows the exploration of the latent space and possibly the creation of new textures. An overview of our method is shown in figure \ref{TAE}.

\vspace*{-.2cm}  

\subsection{Texture encoder}
\begin{figure}[]
    \centering
\includegraphics[width=.80\linewidth]{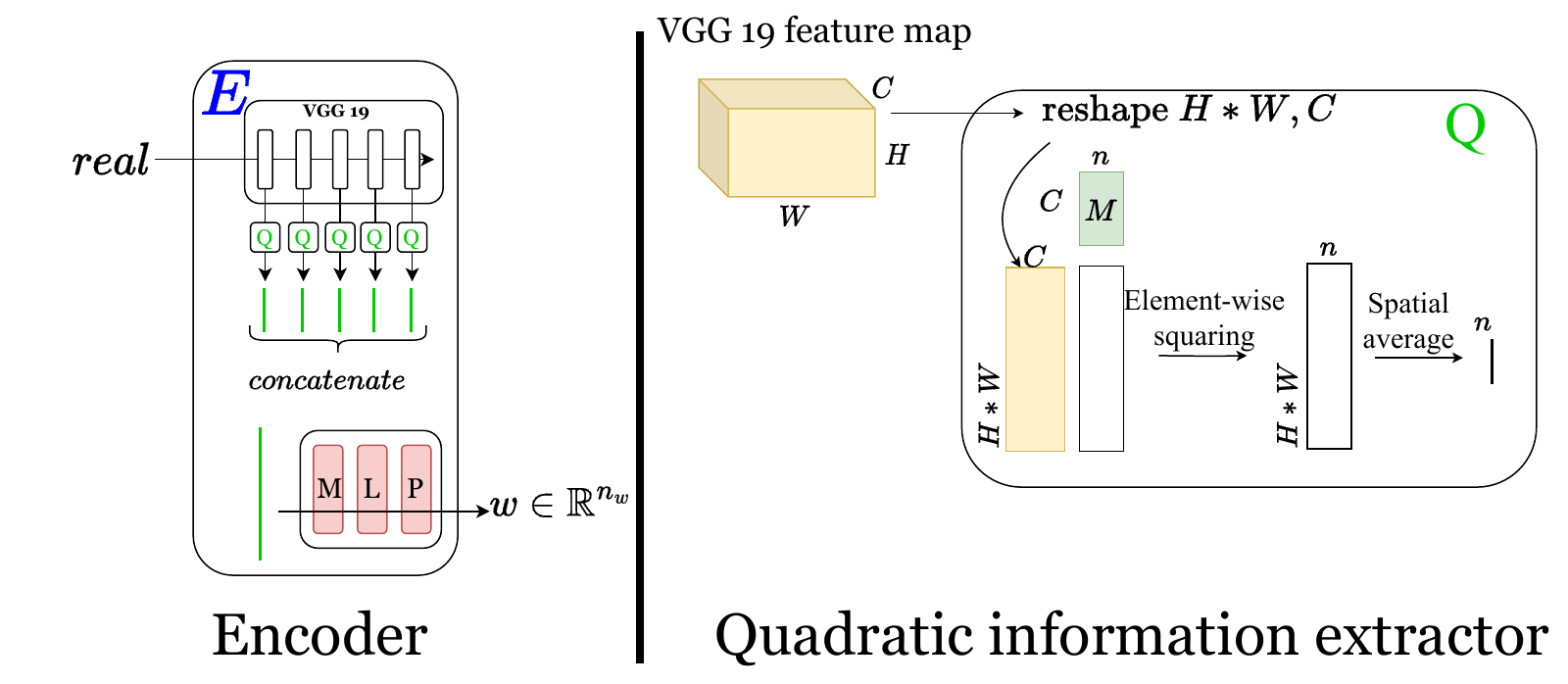}
\caption{Left: Encoder, Right: Quadratic information extraction module}
      \label{quad}
      \vspace*{-.2cm} \end{figure}

In order to develop an encoder embedding texture images in the latent space of StyleGAN \cite{StyleGAN2}, we classically rely on the second order statistics of deep features, in the spirit of \cite{gatys2015texture} and the numerous works that have followed.  Given 5 different depths $l$, we retrieve features $F^l$ from the VGG19 network, each one having dimension $(C_l,H_l W_l)$, to extract their second order statistics. The main difficulty to build a texture encoder from these statistics lies in their high dimensionality. In order to extract a single low dimensional representation from these features, we first extract compact information from the VGG features at each scale (see Fig.\ref{quad}, right), which we then combine into a single compact vector $w=E(I)$ using an MLP. 
At each scale, the second order features are extracted using the technique presented in \cite{second_order_cnn}. Indeed, for each depth $l$, instead of computing whole $(C,C)$ Gram matrices $G^l_{c,c'}=\sum_{k=1}^{H_lW_l} F^l_{c,k} F^l_{c',k}$ as in \cite{gatys} and then reduce their dimensionality, we directly compute from $G$ a representation vector $Q(G)$ of size $n_w$, the chosen dimension of our latent space. These vectors are defined as:
\begin{equation}
    Q(G)_i =m_i^T\cdot G\cdot m_i
\end{equation}
with $\{m_i\}_{1\leq i \leq n}$ a set of trainable vectors of size $(C,1)$. If we write $G=F^l \cdot F^{lT}$, the previous equation becomes:
\begin{equation}
    Q(G)_i =m_i^T\cdot F^l\cdot F^{lT}\cdot m_i =(m_i^T\cdot F^l)\cdot(m_i^T\cdot F^l)^T 
\end{equation}
This allows the extraction of relevant second order features without having to compute the whole Gram matrix (as illustrated in Fig.\ref{quad}), reducing the complexity from $n(2HWC^2+4C) \approx 2nHWC^2$ to $n(2*CHW+2HW)  \approx 2nHWC$,
i.e. by a factor C, which goes up to 512 for the feature maps used in the latest depths of the VGG network.

\vspace*{-.2cm}  \vspace*{-.2cm}

This process results in a compact texture encoder, built only from second order statistics of some learned combination of features and carrying non localized  textural information. This non-localized nature is of course a desirable property in view of texture synthesis and will be shared by the generator. 
\vspace{-0.7em}
\subsection{Texture generator}
We use a modified StyleGAN architecture as our generator G. To perform image synthesis, such an architecture uses a latent variable $w$ in a non-localized way (contrarily to DCGAN-like architectures): $w$ is used to predict and set the first and second moments of feature maps  during the multi-scale synthesis process (through Adaptive Instance Normalization \cite{adain}).

Aiming to remove any systematic spatial information in the generation process, we modify the StyleGAN architecture so that all spatial information is due to the realizations of noise maps at each step of the generation process: instead of learning a tensor of size (512*4*4), we discard spatial dimensions by only learning an input tensor (n*1*1), which we then expand by replication at the start of the generation process. Besides, convolutions use no padding to avoid border artifacts.

Thanks to these changes, our auto-encoder achieves full disentanglement of texture information and pattern localization.
Periodic information is also added at every level of the generator, as will be further described in sections \ref{sec:scale_indep}-\ref{sec:st}.
\vspace{-1em}
\subsection{Sine waves}
In order to synthesize periodic-like textures having long range dependency, we add sine waves to our generator, in a way similar to \cite{PSGAN}, \cite{texton}. Given a set of $n_{freq}$ frequencies $f'_i$  (detailed in \ref{sec:st}),
we build $S\in \mathbb{R}^{n_{freq}*H*W}$ a volume where each channel $S_i$ is a sine wave of frequency $f'_i$: $S=(sin(f'_i\cdot\boldsymbol{x}+\phi))_i,$ with $\boldsymbol{x} \in \mathbb{R}^{H*W}$  the spatial position, and $\phi$  a random phase. 
In contrast to the works in \cite{PSGAN}, \cite{texton}, we modulate elements $S$ in the following ways:
\begin{itemize}
    \item we weight each channel $S_i$ of the sine maps $S$ with the coefficient $weight_{level}(f_i)$ defined in Section \ref{sec:scale_indep}. Its purpose is to inject the frequency $f_i$ in the right level of the generator given its magnitude.
    \item we use the latent variable $w$ to weight the use of every frequency according to the latent representation, using a fully connected layer $A \in \mathcal{M}_{n_{freq},n_w}$, in the spirit of \cite{StyleGAN2}, to project $w$ onto a weighting vector $A\cdot w$.
\end{itemize}
Finally, the addition of periodic content to a given feature map $F \in \mathbb{R}^{B*C*H*W}$ is performed by addition, after applying a 1 by 1 convolution filter to $S$ to reach the number of feature maps $C$ (see Fig.\ref{TAE}, right): 
\begin{equation}
F \longleftarrow F+conv\Big(S *weight_{k_{level}} * A\cdot w\Big)
\end{equation}
\vspace{-.8cm}
\subsection{Scale independent learnable frequencies}\label{sec:scale_indep}
In this section, we describe a novel way to generate periodic information independently from the architecture of the generator.
We define a {\it level} of the network to be the set of operations delimited by a change of resolution. In our case each level is a StyleGAN block, delimited by upsampling by a factor 2.

\begin{figure}[]
    \centering
      \includegraphics[width=1.0\linewidth]{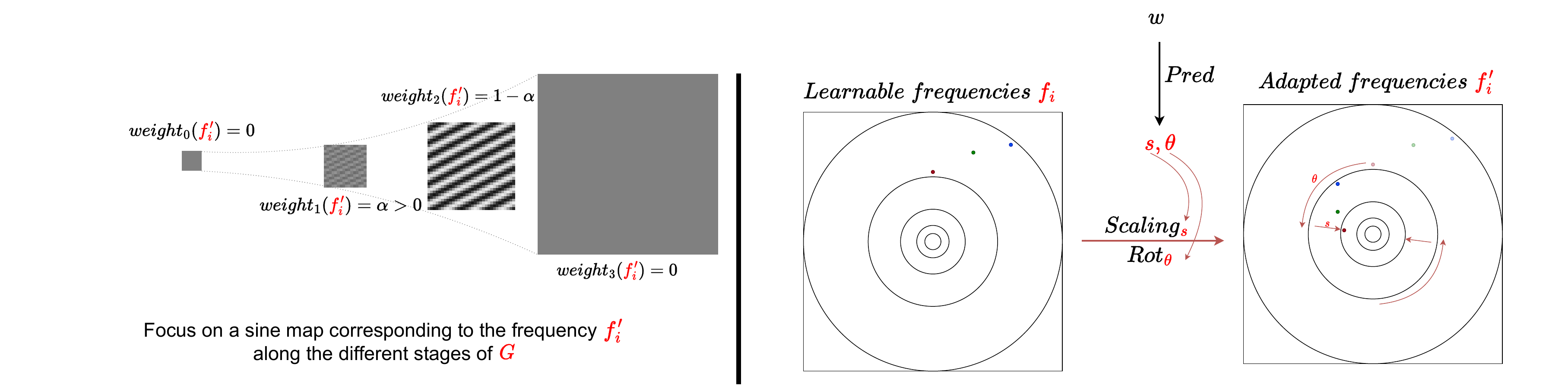}
      \caption{\textbf{Left:} Use of a global frequency at different scales in the generator, depending on its magnitude. \textbf{Right:} Periodic information to incorporate to the network: learnable frequences are transformed using parameters depending on the input (through $w$).}
      \label{freqs}
      \vspace*{-.2cm} \end{figure}

In the literature, periodic content is usually incorporated at specific levels of generators. Sine maps are either used at the lowest resolution level of the network in \cite{PSGAN} (not allowing for high-frequency content to be added), or at each level of the network as in \cite{texton} (leading to sudden discontinuity in the generation process due to the change of level). Instead of learning a different set of frequencies at each level, we choose to learn them independently of the architecture of the generator, and incorporate them in the generator at the adequate levels, depending on their magnitude. The idea is that a frequency $f$ has to be added at the level of the network where its magnitude (relative to the resolution of the level) is not too low nor too high.
The full procedure is illustrated in Fig.\ref{freqs}, left.
      
In short, we learn a set of $n_{freq}$ frequencies $f_i$ independently of the architecture of the generator. Depending on the magnitude of a given frequency, we use it to create sine maps in the two levels of the generator where its magnitude is adequate (relative to the resolution of the level).
The use of a frequency $f$ is then weighted based on $|f|$: for each level number $k_{level}$ in the network , we define the weighting as:
\begin{equation}
weight_{k_{level}}(f)=\begin{cases} 1-\big|1-log_2(\frac{|f|}{f_0\cdot 2^{k_{level}}}) \big| \;\;\;\;\;\; \text{ if }\frac{f_0}{2}\leq \frac{|f|}{ 2^{k_{level}}} \leq  2f_0 \\
0 \text{ otherwise}
\end{cases}       
\end{equation}
where $f_0$ corresponds to a reference magnitude.
This allows for a frequency's use level to smoothly move along the dyadic structure of the generator.

\subsection{Image specific scale and orientation estimation}\label{sec:st}
In this section, we describe a key component of our auto-encoder: each learned frequency $f_i$ is rotated and scaled accordingly to the input, both at training time and for inference. This allows to generate a new sample with appropriate periodic content, aligned with the input. 

For each input image, scale and orientation parameters are learned from the corresponding latent variable $w$.
We directly infer these scale and rotation parameters  with an auxiliary network $Pred$: $Pred(w)=(s,\theta)$.
In order to define a scale and an orientation, one would usually need a reference, with scale $1$ and orientation $0$. We avoid this issue by forcing the network's predictions to be consistent with geometric transformations in the image space. This approach is reminiscent of self supervised learning (SSL) approaches.

Given an input image $I$, we geometrically transform $I$ with two distinct scaling and rotation parameters $\hat{s}_i$ and $\hat{\theta}_i$ and then encode these images to get latent codes:
$w_i=E(Scaling_{\hat{s}_i}(Rot_{\hat{\theta}_i}(I)))$. Finally, we predict transformation parameters from these latent codes: $s_i,\theta_i=Pred(w_i)$.
The corresponding loss $\mathcal{L}_{SLL}$ reads:
\begin{equation}
\begin{split}
    \mathcal{L}_{SSL}=1 - \Big|\frac{\pi}{2} -\big( (\hat{\theta}_1-\hat{\theta}_0) -(\theta_1-\theta_0)\big)mod_{\pi}\Big|
   +\big( \frac{\hat{s}_1}{\hat{s}_0}  - \frac{s_1}{s_0}\big)^2
\end{split}
\end{equation}

The first part of the loss takes into account the $\pi$-periodicity of the orientation of an image and is minimal when the difference in image rotation angles $\hat{\theta}_1-\hat{\theta}_0$ is equal to the difference of predicted angles $\theta_1-\theta_0 \;\; mod\; \pi$. Similarly, the second part of the equation is minimal when the ratio between estimated scales matches the ratio of the scales used in the geometrical augmentation. Eventually, these input-dependent parameters are used to scale and rotate the set of global frequencies $f_i$, effectively adapting the frequency patterns to each specific image, as illustrated in Fig.\ref{freqs}, right. Writing $s_{pred},\theta_{pred}=Pred(w)$, we get: 
\begin{equation}
f'_i=Scaling_{s_{pred}}(Rot_{\theta_{pred}}(f_i))
\end{equation}

\vspace{-1em}
\subsection{Direct sampling}
Using our architecture, the natural way to synthesize a texture is to encode and decode an image in the latent space, performing synthesis by example. Now, it may be desirable to synthesize new textures without exemplar input. In order to do so, we learn the distribution of the encoded training images $E(p_{data})$ in the $\mathcal{W}$ space. We choose an adversarial approach, where a network $T$ is trained to map a noise distribution $p_\mathcal{Z}$ to $E(p_{data})$. We train $T$ jointly with a discriminator $D_\mathcal{W}$, following the WGAN-GP framework \cite{WGANGP}, with a gradient penalty $\mathcal{L}_{GP}$:
\begin{equation}
    \mathcal{L}_{adv}(T)=\mathbb{E}_{z} \big[ D_\mathcal{W}(T(z)) \big]
\end{equation}
\begin{equation}
    \mathcal{L}_{adv}(D_\mathcal{W})=\mathbb{E}_{I\sim p_{data}} \big[ D_\mathcal{W}(E(I))\big]-\mathbb{E}_{z} \big[D_\mathcal{W}(T(z))\big] 
    +\lambda_{GP} \mathcal{L}_{GP}
\end{equation}

\subsection{Losses}

We further constrain the texture synthesis process using the following losses. To compensate for low-frequency artifacts created by the optimization from \cite{gatys2015texture}, we add the spectral loss $\mathcal{L}_{Spe}$ from \cite{NeuralTextureGonthier} (Equation (3.3)) to force the spectrum of the generated image to match the input's spectrum. This loss complements the texture loss, providing necessary information to the network to align the generated periodic content in $I_2=G(E(I_1))$ with the input image $I_1$.

Using sliced histogram matching \cite{sliced_color}, we also implement a color histogram loss between $I_1$ and $I_2$:
\begin{equation}
\mathcal{L}_{Hist}(I_1,I_2)=\mathbb{E}_{x \in \mathbb{R}^3}\sum (sort_x(I_1)-sort_x(I_2))^2
\end{equation}
Here $x$ denotes a randomly sampled color vector along which to perform color histogram matching, and $sort_x$ is the operation that sorts all pixels of an image accordingly to the value of the projection of the pixels's color value onto $x$.
By randomly sampling $x$, we approximate a color histogram distance between $I_1$ and $I_2$.

Our final optimization objective reads:
\begin{equation}
    \mathcal{L}=\mathcal{L}_{style}+ \mathcal{L}_{Hist}+\mathcal{L}_{Spe}+ \mathcal{L}_{adv}+
    \mathcal{L}_{SSL}
\end{equation}


\subsection{Training}
The whole pipeline is trained in an end-to-end fashion, using the Adam optimizer with a learning rate of $10^{-4}$ for 600000 iterations with a batch size of 8. This amounts to a week of training on a single NVIDIA RTX 6000 GPU.

%% file: 04_experiments.tex
\section{Experiments}

\label{sec:exp}
\subsection{Assessed methods, datasets and evaluation metrics}
Among the many texture synthesis algorithms, we choose to compare ourselves to these methods:
\begin{itemize}
    \item TextureCNN \cite{gatys2015texture}, as one of our goals is to shortcut this process with a feed-forward auto-encoder;
    \item PSGAN \cite{PSGAN}, as they introduced the idea of augmenting a generator with periodic content;
    \item Neural texture \cite{neural}, since they also use a texture encoder;
    \item Whitening Coloring Transform (WCT) \cite{WCT}, because this method performs universal texture synthesis with no learning step.
\end{itemize}

We do not compare against \cite{user-controllable} (unavailable code) nor \cite{texton} as no encoder is included in their architecture. For all methods requiring a training (ours, PSGAN and neural texture), we choose a latent space dimension of 32.

PSGAN does not come with an encoder to project an image into a global latent space variable $z_g$. As a baseline, we perform inversion of PSGAN as in \cite{texture_mixer}: given a data sample, randomly initialize a latent variable $z_g$ and perform gradient descent in the latent space to minimize the texture loss $\mathcal{L}_{style}$ between the generated image $G(z_g)$ and the data sample. We keep the best result out of 10 random initializations. We call this resynthesis method PSGAN-GD.

We use the texture loss from \cite{gatys} as a distance to measure performance, along with SIFID (introduced in \cite{SinGAN}) and LPIPS \cite{LPIPS}. We use the dataset MacroTextures introduced in \cite{texton}.
We strongly augment each dataset with geometric transformations: rotations of any angles and scalings to train our method, Neural texture, and PSGAN.
\vspace{-0.9em}
\subsection{Visual and quantitative results}
\begin{figure}[h!]
\centering
\begin{tabular}{cccccc}
\setlength{\tabcolsep}{.1pt}
\renewcommand{\arraystretch}{.1pt}
\centering
GT & TAE (ours) & Universal style & Gatys & Neural Texture  \\

\includegraphics[width=.2\linewidth]{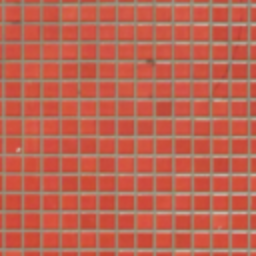}&
\includegraphics[width=.2\linewidth]{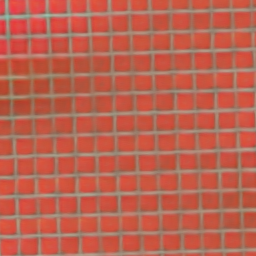}&
\includegraphics[width=.2\linewidth]{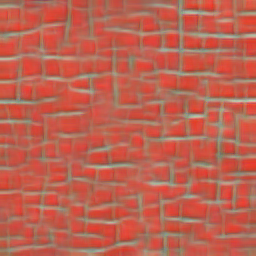}&
\includegraphics[width=.2\linewidth]{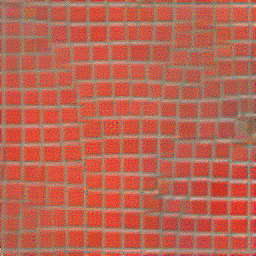}&
\includegraphics[width=.2\linewidth]{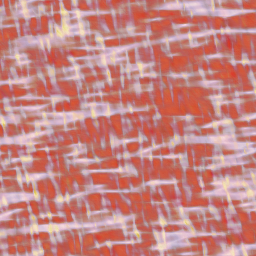} \\

\includegraphics[width=.2\linewidth]{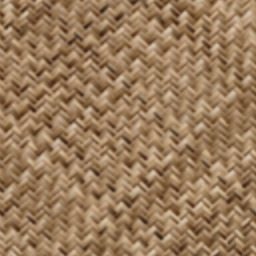}&
\includegraphics[width=.2\linewidth]{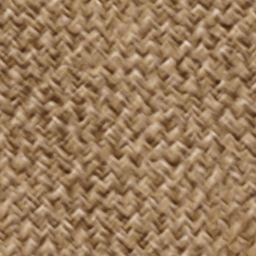}&
\includegraphics[width=.2\linewidth]{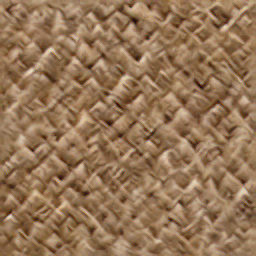}&
\includegraphics[width=.2\linewidth]{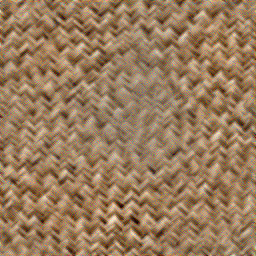}&
\includegraphics[width=.2\linewidth]{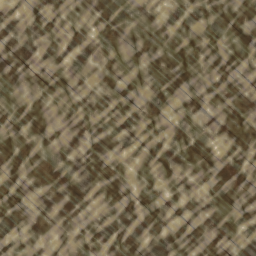} \\

\includegraphics[width=.2\linewidth]{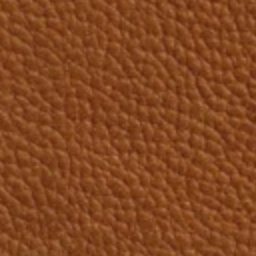}&
\includegraphics[width=.2\linewidth]{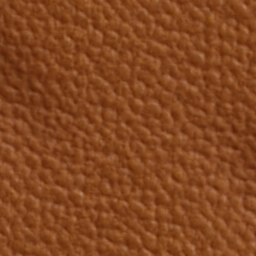}&
\includegraphics[width=.2\linewidth]{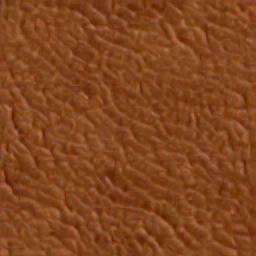}&
\includegraphics[width=.2\linewidth]{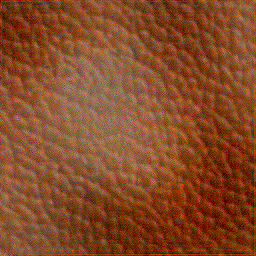}&
\includegraphics[width=.2\linewidth]{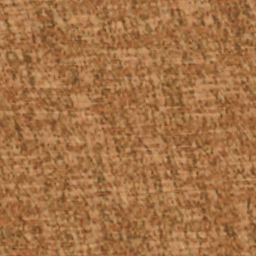} \\

\includegraphics[width=.2\linewidth]{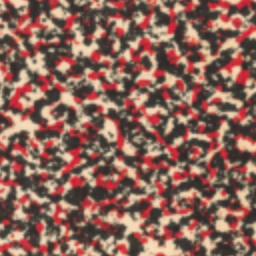}&
\includegraphics[width=.2\linewidth]{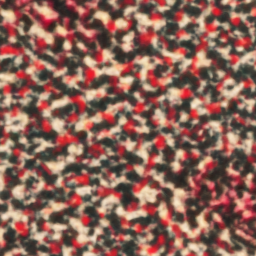}&
\includegraphics[width=.2\linewidth]{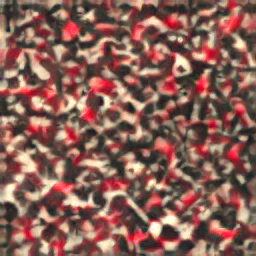}&
\includegraphics[width=.2\linewidth]{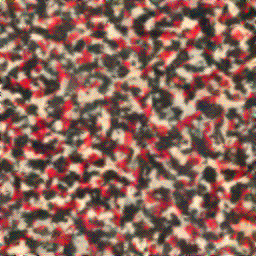}&
\includegraphics[width=.2\linewidth]{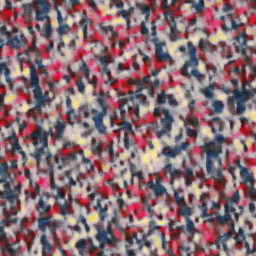} \\

\includegraphics[width=.2\linewidth]{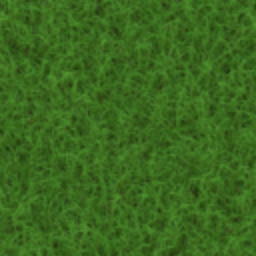}&
\includegraphics[width=.2\linewidth]{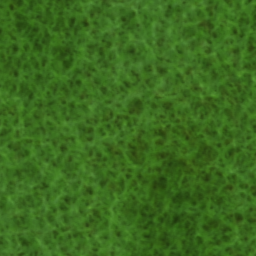}&
\includegraphics[width=.2\linewidth]{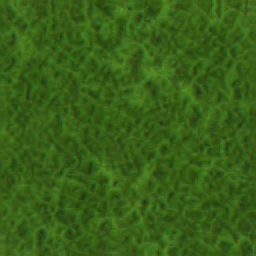}&
\includegraphics[width=.2\linewidth]{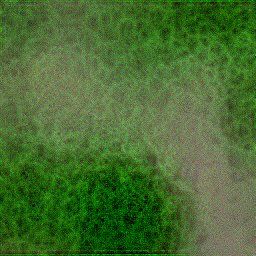}&
\includegraphics[width=.2\linewidth]{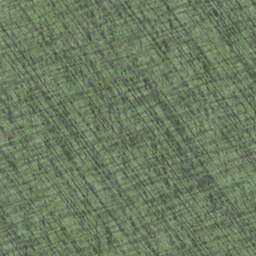} \\

\end{tabular}
\caption{Comparison of syntheses obtained from 4 different methods.}
\label{comp_fig}
      \vspace*{-.2cm} \end{figure}

Synthesis results are simply obtained by first embedding an input exemplar into the latent space (using the encoder) and then generating a new sample from this latent code. Variability of the synthesized samples is classically obtained thanks to the noise maps in the generator. A comparison of syntheses obtained with the different methods is presented in Fig. \ref{comp_fig}. We observe that we indeed maintain appropriate long range dependencies, which is not the case in for TextureCNN nor WCT. We also get rid of low frequency artifacts mostly present in TextureCNN.

We show quantitative evaluations in Table 1. Our approach is very close to TextureCNN for the style metric, which leads to the best result as it directly optimizes $\mathcal{L}_{style}$ over the image.

Our method comes second best, notably beating WCT by a relatively big margin. Additionally, our method is orders of magnitude faster. We also obtain the best performance for the SIFID and LPIPS metrics, in particular outperforming the original TextureCNN approach. The results of PSGAN-GD are poor, which is essentially due to the instability of the inversion procedure in the latent space. 

\newcommand{\figa}{9}
\newcommand{\figb}{6}
\newcommand{\figc}{15}
\newcommand{\figd}{29}
\newcommand{\fige}{35}
\newcommand{\figf}{6}
\newcommand{\figg}{12}

\vspace*{-.2cm}  \vspace*{-.2cm}

\begin{table}[h!]
\captionsetup{justification=centering}\caption{Evaluation of the methods using $\mathcal{L}_{style}$, SIFID, LPIPS and runtime in seconds (all on one NVIDA RTX6000 GPU) on the MacroTextures dataset}
\centering
\scalebox{1}{
 \begin{tabular}{|l|c|c|c|c|} 
 \hline
 & \;\;\;\;$\mathcal{L}_{style}$ ($\cdot 10^{3}$) \;\;\;\;&\;\; SIFID ($\cdot 10^{-6}$) \;\;&\;\;\;\;\; LPIPS \;\;\;\;\;&\;\;\;\; runtime\;\;\;\;\\ 
  \hline
  TextureCNN \cite{gatys2015texture}& $ \bold{3.8 \pm 16.8}  $ & $  14 \pm 14  $ & $  0.51 \pm 0.09 $ & $ 60$ \\ \hline
  
  TAE (ours)  & $ 5.2 \pm 11.7 $ & $ \bold{9 \pm  11}$ & $ \bold{0.35 \pm 0.09} $ & $0.085$ \\ \hline
  
  Neural texture \cite{neural}& $ 135 \pm 468 $ & $ 178 \pm 190 $ & $ 0.56 \pm 0.09 $ &  $\bold{0.0072}$ \\ \hline 
  WCT \cite{WCT} & $ 29.8 \pm 132.9 $ & $ 10 \pm 18 $ & $ 0.44 \pm 0.07 $ & $ 3.5 $ \\ \hline
  
  PSGAN-GD \cite{PSGAN}& $ 208 \pm 611 $ & $ 381\pm 607 $ & $ 0.64 \pm 0.11 $ & $ 24$\\ \hline
 \end{tabular}}
 \label{table1}
 \vspace*{-.4cm}  
\end{table}

\subsection{Geometric completeness}
\begin{figure}[h!]
    \centering
      \includegraphics[width=.90\linewidth]{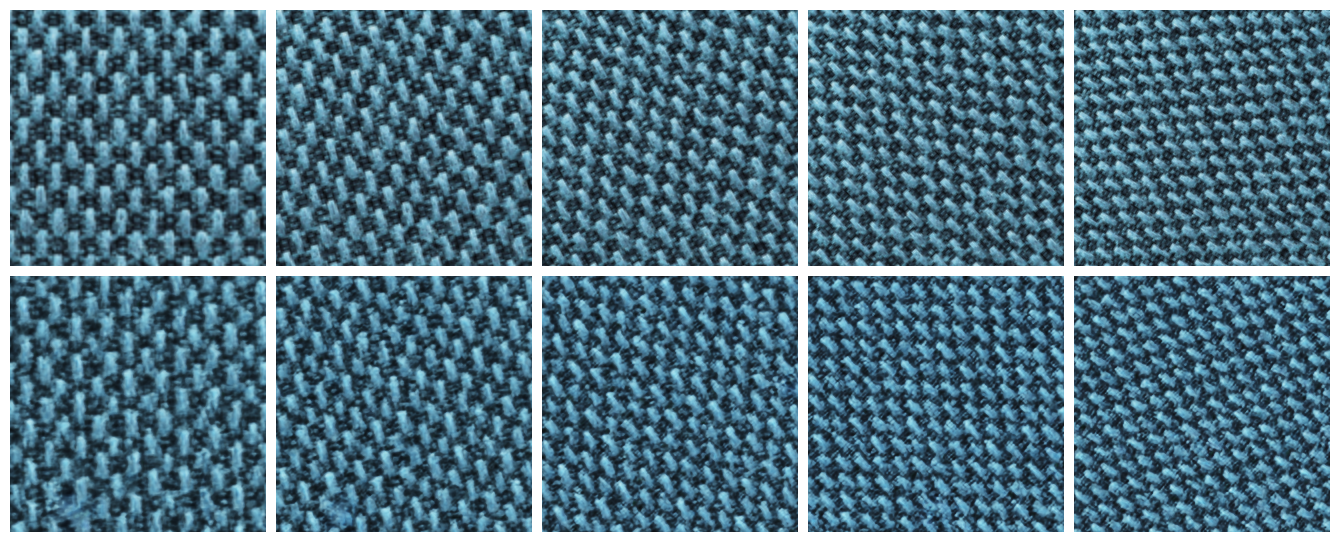}
      \caption{Top row: texture image geometrically augmented, Bottom row: feed forward texture resynthesis $G(E(I))$}
      \label{augmentation}       \vspace*{-.2cm} \end{figure}

In order to illustrate the ability of the method to deal with simple geometric transforms, we perform auto-encoding on various scalings and rotations of the same input image, thus showing successful adaptation to the input in Fig.\ref{augmentation}.

      \begin{figure}[h!]
    \centering
    \includegraphics[width=.8\linewidth]{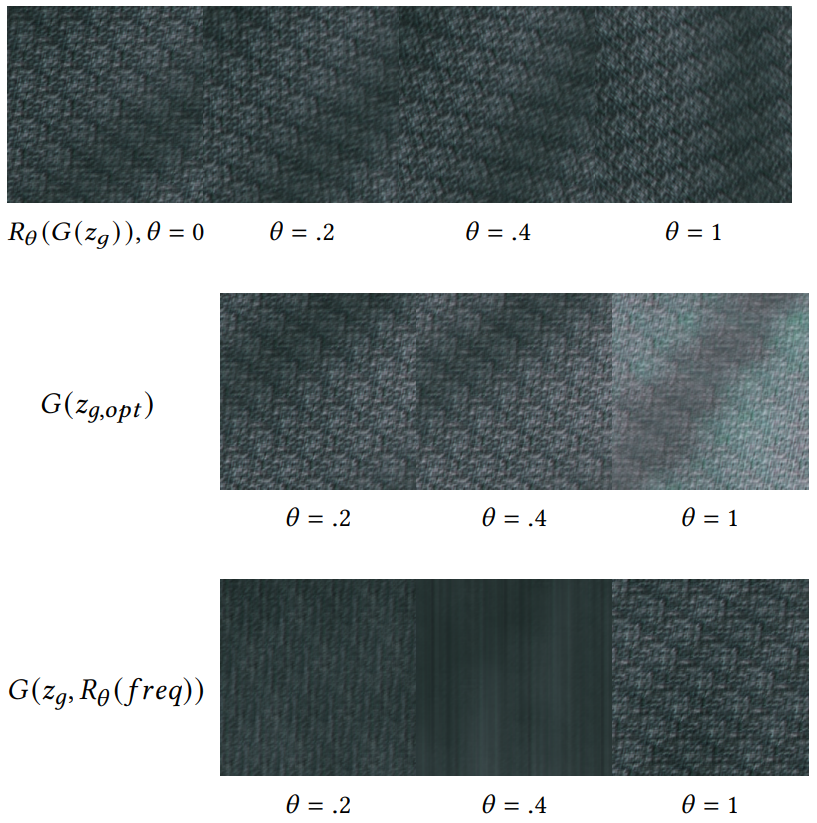}
\caption{Illustration of the difficulty of PSGAN to deal with arbitrary rotations \newline \textbf{First row}: Rotations of input texture $R_\theta(I)$, $I=G(z_g)$ - \textbf{Second row}: PSGAN samples where $z_{g,opt}$ is obtained by gradient descent to match the texture $R_\theta(I)$ - \textbf{Third row}: Directly rotating the frequencies without changing $z_{g}$}
    \label{PSGAN_rot}
      \vspace*{-.2cm} \end{figure}
      
As a comparison we show that PSGAN, and thus the methods based upon it (e.g. \cite{user-controllable}), lack this ability.  To exhibit that PSGAN indeed does not have the capacity of handling different rotations of the same textures, we start by sampling a global variable $z_g$, generating an image $I=G(z_g)$.

Then, we create different rotations of $I$: $R_\theta(I)$ as shown in the first row of Fig.\ref{PSGAN_rot}, and try to reproduce them with the generator $G$. Indeed, taking $I=G(z_g)$ avoids inversion problems. On the contrary, as observed in Fig. 6a) of \cite{user-controllable}, the latent space of PSGAN is not smooth, thus the inversion of this network is 
unstable. 

The second row shows $G(z_{g,opt})$, where $z_{g,opt}$ is obtained by minimizing  $\mathcal{L}_{style}(R_\theta(I),G(z_{g,opt}))$ gradient descent. We notice that the periodic structure is not rotated, while smaller details are correctly oriented. This is indeed not a satisfactory result. In the PSGAN architecture, the frequencies of sine maps are inferred from $z_g$. We also tried to directly rotate these frequencies with the same angle $\theta$ as we rotated the image $I$ with, thus aligning the periodic content with the image we want to reconstruct, $R_\theta(I)$. This operation yields poor pattern fidelity albeit somewhat aligned, as can be seen on the third row.

These two tests prove that, although trained on a dataset augmented with rotations, locally in the latent space of PSGAN, each texture is intrinsically binded to an orientation. There is no way to change the orientation without harming texture fidelity.
\captionsetup[subfloat]{labelformat=empty}

\vspace{-1em}
\subsection{Spatial interpolation}
\begin{figure}[h!]
    \centering
    \includegraphics[width=.46\linewidth]{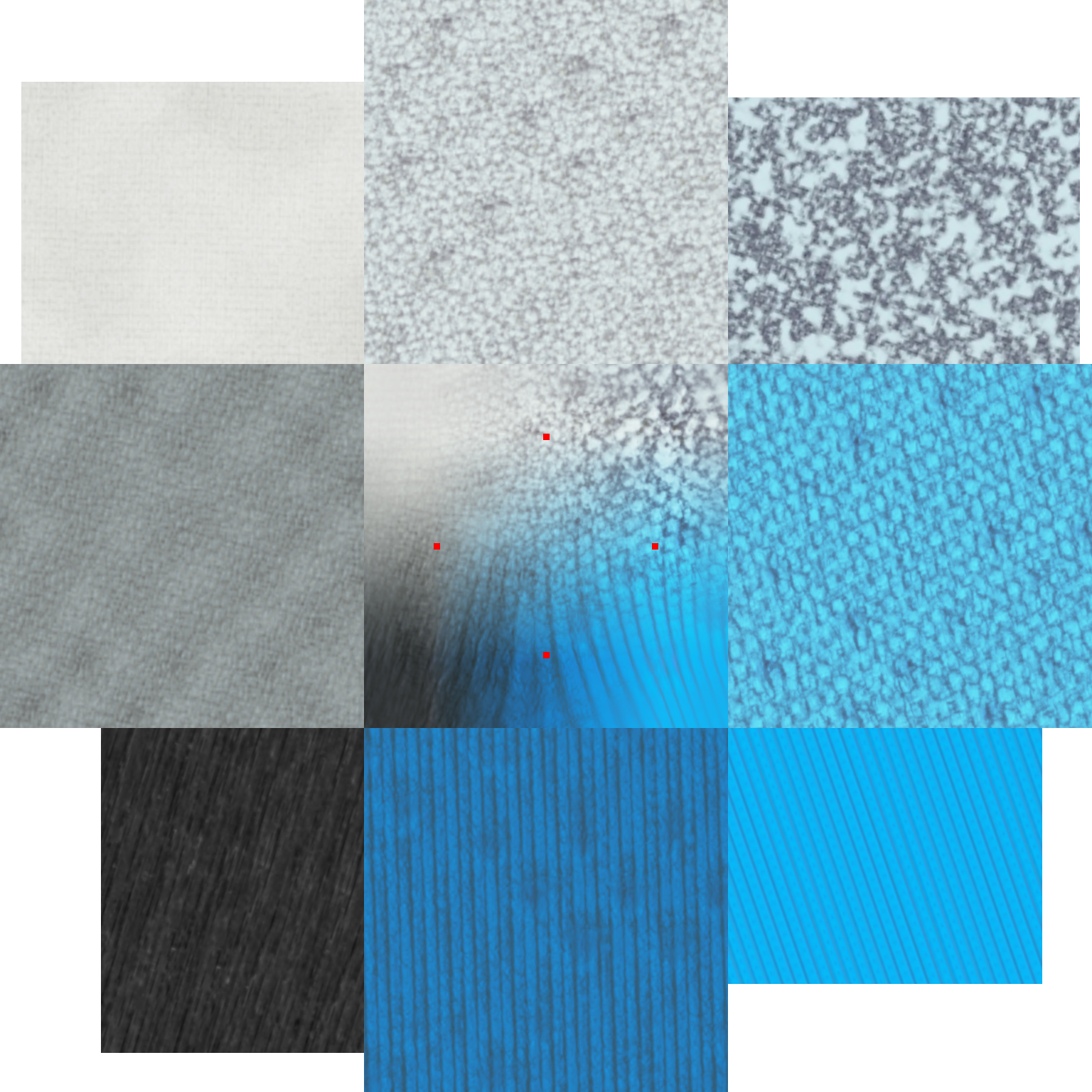}\hfill
      \includegraphics[width=.18\linewidth]{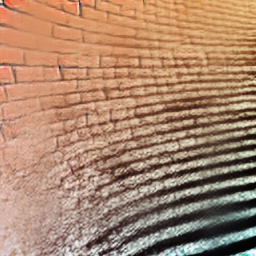}\includegraphics[width=.36\linewidth]{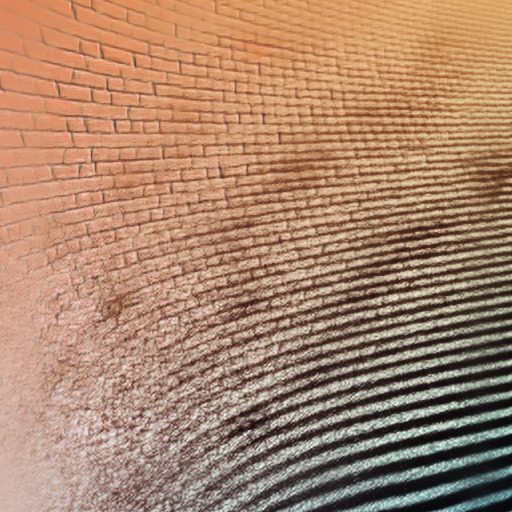}
      \caption{\textbf{Left:} Texture Palette. Four input textures of various sizes  and aspect ratio are displayed in the corners. In the center, an image shows a field obtained by performing spatial interpolation between the corner textures. On this field, the red dots are positions from which the images in the middle of each side are synthesized.
      \newline \textbf{Right:} Interpolation between 4 textures without and with expansion by a factor 2}
      \label{interpolation}
      \vspace*{-.2cm} \end{figure} \vspace*{-.2cm} \
Our approach relies on a texture latent space and is therefore naturally adapted to texture interpolation, for instance by simply linearly interpolating latent variables. Spatial interpolation (building an image across which a texture is progressively varying), on the other hand, is more challenging. Indeed, AdaIN layers normalize features based on statistics computed across the whole image, therefore globally defining new features. We solve this problem by computing statistics locally. Additionally, the mean and variance imposed after normalization are computed from a texture representation $w$ varying in the spatial domain, to allow a smooth transition of the content. An asset of this method is that the periodic content has a natural variation in space. We can also use it to create interpolated texture from a visual palette (Fig.\ref{interpolation}, left). We show an example of texture interpolation and of interpolation combined with expansion in Fig.\ref{interpolation}, right.


%% file: 10_conclusion.tex
\section{Conclusion}
\label{sec:conclusion}
We introduced a feed forward auto-encoder network having the ability to learn a set of textures, yielding fast exemplar-based synthesis results.
Thanks to adaptive periodic content, our method allows texture encoding and reconstruction of any orientation and scale. Its disentanglement of texture characteristics and spatial distribution allow for common texture manipulations such as interpolation and expansion. The proposed approach outperforms closely related methods for usual texture fidelity metrics.